\documentclass[letterpaper]{article}

\usepackage[utf8]{inputenc}
\usepackage{natbib,alifeconf}
\usepackage{amsmath}
\usepackage{amssymb}
\usepackage{url,hyperref,cleveref}
\usepackage{booktabs}

\hypersetup{
    colorlinks=true,
    citecolor=red,
    }

\usepackage{caption}
\usepackage{subcaption}

\newcommand\blfootnote[1]{%
  \begingroup
  \renewcommand\thefootnote{}\footnote{#1}%
  \addtocounter{footnote}{-1}%
  \endgroup
}

\title{Distilling a Modular Reservoir Through a Genomic Bottleneck}

\author{
    Mani Hamidi$^{1, 2,*}$,
    Sina Khajehabdollahi$^{3}$,
    Charley M. Wu$^{2,4,5, +}$,
    Emmanouil Giannakakis$^{1,2, 6, +}$,
    \mbox{}\\
    $^1$Department of Computer Science, University of T\"ubingen, T\"ubingen, Germany \\
    $^2$Max Planck Institute for Biological Cybernetics, T\"ubingen, Germany \\
    $^3$Flowers Team Inria, Bordeaux, France \\
    $^4$ Human and Machine Cognition Lab, Center for Cognitive Science, TU Darmstadt, Darmstadt, Germany \\
    $^5$ Hessian.AI, Darmstadt, Germany \\
    $^6$Department of Bioengineering, Imperial College London, London, UK \\
    $^+$Equal contribution\\
    $^*$mani.hamidi@uni-tuebingen.de \\ \hspace{2cm}
} 

\begin{document}

\maketitle
\begin{abstract} 
The intricate structures of biological neural networks largely emerge during development, guided by a comparatively compressed blueprint encoded in the genome. The connectivity that emerges from this decoding process is rich in structure, and already equips the organism with functional modules upon birth. This initial structure serves as a scaffold that can be gradually refined and fine-tuned through lifelong experience, via a variety of plasticity mechanisms. 
Drawing inspiration from this interaction between evolutionary and developmental modes of learning, we use hypernetworks to learn a compressed generative process that generates the connectivity of a modular reservoir. 
We show that this marriage between curriculum-based meta-learning and
modular reservoir computing can generate
sparse recurrent networks that solve difficult temporal tasks with minimal training and without concessions to robustness.

\end{abstract}

Data/Code available at: \url{https://github.com/manihamidi/evo-devo-reservoir-prod}

\section{Introduction}

The architecture\blfootnote{\textcopyright\ 2026 Mani Hamidi, Sina Khajehabdollahi, Charley M. Wu, and Emmanouil Giannakakis. Published under a Creative Commons Attribution 4.0 International (CC BY 4.0) license.} of biological neural networks is determined by a combination of \textit{genetically} encoded mechanisms---refined over evolutionary timescales and expressed early in development---and adaptive \textit{phenotypic} changes driven by neuroplasticity that modify connectivity within an organism’s lifetime \citep{eschbach2021circuits}. %
Genetically encoded mechanisms such as axonal guidance \citep{stoeckli2018understanding} already give rise to a highly structured and modular scaffolding during early development and embryogenesis. Such genetically encoded modularity can aide learning both over evolutionary time scales as well as during the lifetime of the organism \cite{meunier2010modular}. Evolution can duplicate and reuse the genetic programs that have already proven as functionally efficacious units \cite{Stephenson-Jones2011-qy, Grillner2016-yu}, and lifelong learning can simply refine already functional networks via experience, rather than reinvent them from scratch.

Biological systems therefore split the burden of learning across both genetic modifications over evolutionary timescales, as well as phenotypic modifications during the lifetime of an agent. Evolution distills powerful inductive biases into compressed genomic representations that act as an input to the generative mechanisms responsible for development and embryogenesis \citep{Mitchell2025-dr}. Thus, rather than starting from a blank slate, organisms benefit from the informational content of both the blueprint programmed into the genome and the self-organizational process of development \citep{Collinet2021-qc}, leading to an information-rich phenotypic product. %
Therefore, the evolutionary-developmental process --- leveraging genetic encoding and developmental decoding --- can be considered as a form of ``genomic bottleneck`` \citep{shuvaev2024encoding} that induces compression of the final phenotypic product onto the genetic substrate. This compression mechanism is not only ``lossless`` but has been theorized to be critical for flexible, robust, and generalizable intelligence \citep{Mitchell2025-dr, igl2019generalization, zador2019critique, kleinman2023cortical, Crosscombe2023-jf}.

For decades, the intricacies of evolutionary-developmental principles have been extensively studied in a subdiscipline of biology \citep[``evo-devo``;][]{carroll2013dna}. A growing body of recent work suggests that the application of the same principles, from compositional pattern-producing networks \citep[CPPNs;][]{stanley2007compositional}, genomic bottlenecks  \citep{shuvaev2024encoding, Barabasi2023-ok} and neural development programs \citep[NDPs;][]{najarro2023towards, nisioti2024growing, pedersen2024structurally, plantec2024evolving}, can also benefit artificial systems.
Specifically, meta-learning methods that generate structural priors for artificial networks \citep{finn2017model, liu2018darts, shaw2019meta, elsken2020meta, Ha2016-vz, shuvaev2024encoding} have been shown to maintain or enhance network performance compared to classical, over-parameterized networks trained \emph{de novo} to solve a given task.

Another consequence of encoding phenotypic structure into genetic representations is the relative ease with which the same modular units can be duplicated and reused thanks to gene duplication events \citep{Garcia-Fernandez2005-lj, Grillner2016-yu}.
The duplication of core modular units over evolutionary timescales allows for the same unit to be co-opted through the process of ``exaptation'' and extend the organism's functional repertoire to new and more complex behavioral traits \citep{Stephenson-Jones2011-qy, Cisek2019-cn}. 
Indeed, modular structures are ubiquitous in biological neural networks \citep{bertolero2015modular, meunier2010modular, casanova2019modular}. From a theoretical standpoint too, modularity has been shown to act as a strong inductive bias, helping achieve higher performance and trainability in biological networks \citep{zhang2024inductive, clune2013evolutionary}, while also significantly boosting performance and efficiency in artificial networks \citep{rodriguez2019optimal, tahmasebi2012fast, Hamidi2024-hx}.

\begin{figure}[t]
    \centering
\includegraphics[width=1.0\columnwidth]{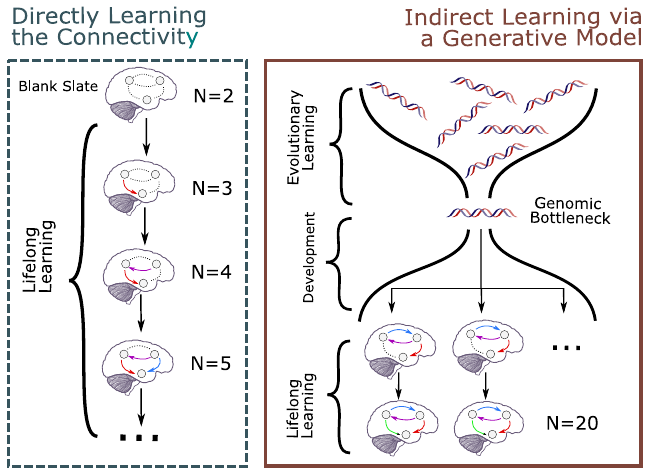}
    \caption{\textit{\textbf{Learning Connectivity Directly via Experience vs. Indirectly via the Genome.} In biological systems learning unfolds both within an individual’s lifetime and through genetic inheritance shaped by a much longer evolutionary process, the latter being encoded on a highly compressed substrate that has passed through a genomic bottleneck. The wiring of brain connectivity during early development, prior to any experience-dependent learning, offers a very powerful inductive bias that can drive efficient lifelong learning during an agent's lifetime.}}
    \label{fig:conceptual}
    \vspace{-1em}
\end{figure}

The reservoir computing framework offers an ideal computational paradigm to capture both modularity and evo-devo. Reservoir computing utilizes a learning paradigm where the input is first projected onto a high-dimensional and complex dynamical module (the ``reservoir``), before being decoded using only a simple linear projection \citep{Tanaka2019-us, lukovsevivcius2009reservoir, Gallicchio2021, Czegel2021-qs}. Crucially, the parameters of the reservoir are predefined and fixed throughout the learning process, leaving only the linear readout to be learned. The procedure therefore lends itself to the ``evo-devo`` learning scheme (Figure~\ref{fig:conceptual} right panel) where the fixed connectivity of the reservoir are first learned during evolution (``evo''), and expressed during embryogenesis and development (``devo'') to give rise to largely functional modules that merely have to be fine-tuned or interfaced by read-out connections. 

Drawing on these biological and theoretical insights, we construct a hierarchical recurrent neural network consisting of a modular reservoir whose connectivity is generated by a highly compressed mechanism---akin to the evolutionarily honed genetic programs that wire the brain during development.
Here, we use ``g-net'' to refer to a hypernetwork that learns a genetic encoding, and ``p-net'' to refer to the phenotypic product of the decoding process that performs the task \citep{shuvaev2024encoding}. 
Using the same working-memory benchmarks---$N$-parity and $N$-delayed match-to-sample---on which the hierarchical modular architecture was previously demonstrated \citep{Hamidi2024-hx}, we show how training the g-net to generate the connectivity between modular reservoirs can learn to solve tasks of equally high complexity but using orders of magnitude fewer trainable parameters than learning the connectivity of the p-net directly. The number of parameters also shows much more favorable scaling properties with respect to both task complexity ($N$) and module size. The latter, in particular, allows our indirect g-net-mediated learning approach to efficiently generate networks that are highly robust to perturbations despite having fewer parameters.

\section{Methods}

\subsection{Tasks}

We evaluate the models on two tasks that are commonly used as benchmark for testing working-memory capacity in recurrent networks; the $N$-DMS (Delayed Match to Sample) and the $N$-parity task. For both, the input is a binary sequence of arbitrary length, presented one bit per time step. For $N$-DMS, the network must output whether the last presented digit and the digits presented in the previous $n$ steps (where $1 \le n \le N$ ) match. For $N$-parity, the network outputs the parity (binary sums) of the last $n$ bits for  $1 \le n \le N$. Thus, a single trained model is expected to solve every parity sub-task up to length $N$. As $N$ grows, the network must maintain more digits in memory, making the tasks progressively more difficult. 

All qualitative findings---the parameter efficiency of the g-net, its robustness advantage under perturbation, and the compressibility of the generated weights---hold for both tasks (Figures~\ref{fig:performance}--\ref{fig:svd_within}). Since $N$-parity is the more demanding benchmark, we use it as the primary illustration for panels that show individual network examples (e.g., weight heatmaps and traces), while quantitative comparisons include both tasks throughout.

\subsection{Structure of the p-net}
\begin{figure}[t]
    \centering
\includegraphics[width=0.99\columnwidth]{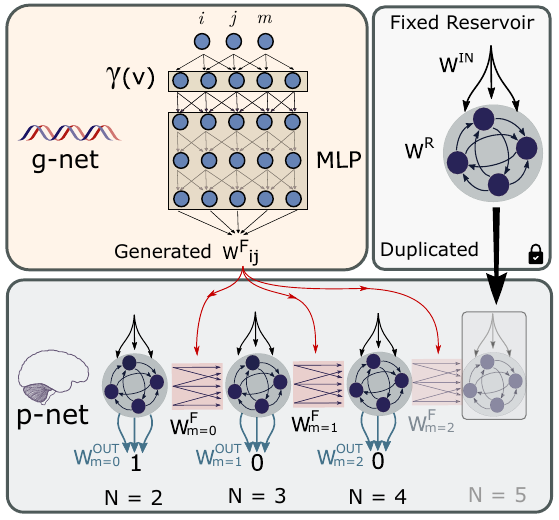}
    \caption{\textit{\textbf{Implementing an indirect learning scheme using a g-net.} Learning takes place at two levels: in the ``evolutionary" or genomic level, the hypernetworks (``g-nets''), are trained to generate just the inter-module weights of the RNN (the ``p-net''). In the ``lifelong learning" phase, the remaining input and output parameters of the p-net are further trained. b) following past \citep{khajehabdollahi2024emergent, Hamidi2024-hx} work we train this g-net following a curriculum}}
    \label{fig:methods}
    \vspace{-1em}
\end{figure}
The phenotypic network or p-net that solves the $N$-parity task follows the growing hierarchical modular architecture proposed by \cite{Hamidi2024-hx}. %
The network consists of multiple modular units each with the same input, $\mathbf{W}^{\text{IN}}$ and recurrent, $\mathbf{W}^{R}$ connections (See Figure~\ref{fig:methods}). The growth of the network adheres to a curriculum  whereby a new copy of the module is added to solve increasingly difficult tasks. Unlike $\mathbf{W}^R$ and $\mathbf{W}^{\text{IN}}$, the feedforward connections, $\mathbf{W}^F$ that link the newly added module to the last are continuously trained throughout the learning process and therefor vary from one module to the next.

More specifically, a single recurrent module with 20 to 80 neurons is trained to solve $N=2$. Once it reaches a performance threshold (accuracy $>98\%$), the recurrent and input connections are duplicated and fixed to be reused by the new module, without any further training. The old module continues solving the task for $N=2$, while the new module is assigned to solve $N=3$. Once the desired accuracy is reached for both, another duplicate copy of the module is added and the same procedure repeats.
The recurrent activation of neurons, $r_i(t)$, in the module $m$ can be summarized as follows:  

\begin{equation}
\begin{aligned}
r_i(t + 1) ={} & \Bigg[ \sum_{ j \neq i}^{M} W^R_{ij} \cdot r_j^m(t) \\
+ & \sum_k^{M} W^{F}_{ik}(m) \cdot r_k^{m-1}(t) \\
+ & \; W^{\text{IN}}_{i} \cdot S(t) + b_i\Bigg]_{\alpha
}
\end{aligned}
\end{equation}

\noindent Where $[\cdot]_{\alpha}$ indicates a leaky ReLU function with negative slope $\alpha$. $\mathbf{W}^{\text{IN}}$ are the input connections, $\mathbf{W}^R$ the in-module recurrent connections, $W^{F}_{ik}(m)$ the weight of the feedforward connection from node $i$ in module $m-1$ to node $k$ in module $m$, $S(t)$ is the binary input signal and $b_i$ is a bias term. $M$ indicates the number of neurons in each module.
There is a separate readout head for each task $N$, linearly mapping the $M$ activations of the module $m=N-2$ into two neurons each corresponding to one of the two parity classes. These linear transformations $\mathbf{W}^{\text{OUT}}_{m}$ are learned via backpropagation throughout training. A greedy policy is used to select the class and compute accuracy with a cross-entropy loss on the output neurons. 

Note that by fixing $\mathbf{W}^F_m$ in addition to $\mathbf{W}^{\text{IN}}$ and $\mathbf{W}^R$, we leave only $\mathbf{W}^{\text{OUT}}_{m}$ for training thus yielding a modular reservoir computer. However, the duplicate-and-fix strategy used for $\mathbf{W}^{\text{IN}}$ and $\mathbf{W}^R$, was already proven untenable for $\mathbf{W}^F_m$ by \cite{Hamidi2024-hx}, prompting us to resort to an alternative strategy to learn them. 

For both tasks, we used module sizes $M \in \{20, 40, 60, 80\}$, with 8 independent runs per combination of training protocol and module size.                     
   
\subsection{Direct vs. Indirect Training of Feedforward Connectivity}

We now turn to the two learning schemes we use to train the feedforward connections, $\mathbf{W}^F_m$, corresponding to the two strategies introduced in Figure~\ref{fig:conceptual}. The first protocol \textit{directly} trains $\mathbf{W}^F_m$ via backpropagation, analogous to phenotypic modification via neuro-plasticity during an organism's lifetime. The second protocol \textit{indirectly} learns $\mathbf{W}^F_m$ by training a generative model (the g-net) whose outputs produce them, analogous to evolutionary modifications to genomic mechanisms that are expressed during development.

\subsubsection{Lifelong learning, directly via the p-net.}

In the case of direct learning during the agent's lifetime, the value of the feedforward weight is modified using the error signal back-propagated from the network's output for all current tasks $N=[2, 3, ..., N_{\max}]$. Specifically, upon solving $N=[2]$, the a copy of the reservoir module is appended to the p-net which at this point still only consists of a single module. The $\mathbf{W}^F_{m=1}$ values are initialized at random to connect the $M$ neurons in the first ($m=0$), module to the $M$ neurons in the second module ($m=1$). The values of $\mathbf{W}^{\text{IN}}$ and $\mathbf{W}^{R}$ in this second module are duplicates of those in the first module and remain fixed throughout training, as described before. After this growth step, the p-net is now trained for tasks $N=[2,3]$, until it achieves $>98\%$ accuracy on both. The procedure repeats as the network grows to solve larger $N_{\max}$, as both $\mathbf{W}^{F}_m$ and $\mathbf{W}^{\text{OUT}}_m$ are modified by the backpropagation signal. 

\subsubsection{Evolutionary learning, indirectly via the g-net.}
In the alternative training method, the feedforward connections between modules in the p-net are not directly modified by the backprop signal but are instead indirectly \textit{generated} by a separate network, the g-net, whose parameters and architecture determine the values of $\mathbf{W}^{F}_m$. The g-net is a hypernetwork \citep{Ha2016-vz} with a simple MLP architecture with three layers of 16 neurons each. The input to the g-net is the index of the source neuron $i$ in module $m-1$, the index of the target neuron $j$ in module $m$, and the module index $m$. Given these three inputs, the g-net, generates corresponding feed-forward weights:
\begin{equation}
    W^{F}_{ij}(m) = g_{\mathbf{\Theta}}(i, j, m).
\end{equation}
\noindent Specifically, upon solving the task $N=2$, instead of randomly initializing  the feedforward connectivity to the second module ($\mathbf{W}^{F}_{m=1}$) and modifying them using backprop, we generate $W^{F}_{ij}(m)$  for all $i, j \in [1, ..., M]$ and $m=1$ using multiple forward passes through the g-net. The error signal is then backpropagated to the parameters of the g-net, $\mathbf{\Theta}$, which will indirectly learn the appropriate $\mathbf{W}^{F}_{m=1}$. 
In practice, these forward passes are batched into a single pass without batch-normalization. 
At later stages in the curriculum when the network is being tested on $N = [2, 3, ..., N_{\max}]$, more forward passes are required to generate $\mathbf{W}^{F}_m$ for all $m \in [1, ..., N_{\max}-1]$

Finally, the index values $v=\{i, j, m\}$ are not used as integer inputs but are each embedded into dense vector representations (indicated as $\gamma(v)$ in Figure~\ref{fig:methods}), typical for categorical data.
These embeddings are learned along with the other weights and biases of the MLP as part of the end-to-end backpropagation scheme, collectively indicated by $\mathbf{\Theta}$. %

\subsection{Perturbation Experiments}
For all perturbation experiments, we sampled the perturbation to each weight in $\mathbf{W}^{F}_m$, from a Normal distribution $\Delta_{ij} \sim \mathcal{N}(0,1)$ but modulated their magnitudes using a scaling factor $\varepsilon$:
\begin{equation}\label{eq:perturb}
    \mathbf{\widetilde{W}}^F_m = \mathbf{W}^F_m + \varepsilon \frac{\mathbf{\Delta}}{||\mathbf{\Delta}||}
\end{equation}
A range of different scaling factors, $\varepsilon \in [1,..., 12]$ were used to examine the impact of different perturbation magnitudes. Notably, perturbations are not scaled by $||\mathbf{W}^{F}||$, so that the same $\varepsilon$ applies an identical absolute perturbation to both training protocols, whose weights differ in scale and variance.

\section{Results}

\begin{figure}[t]
    \centering
\includegraphics[width=0.99\columnwidth]{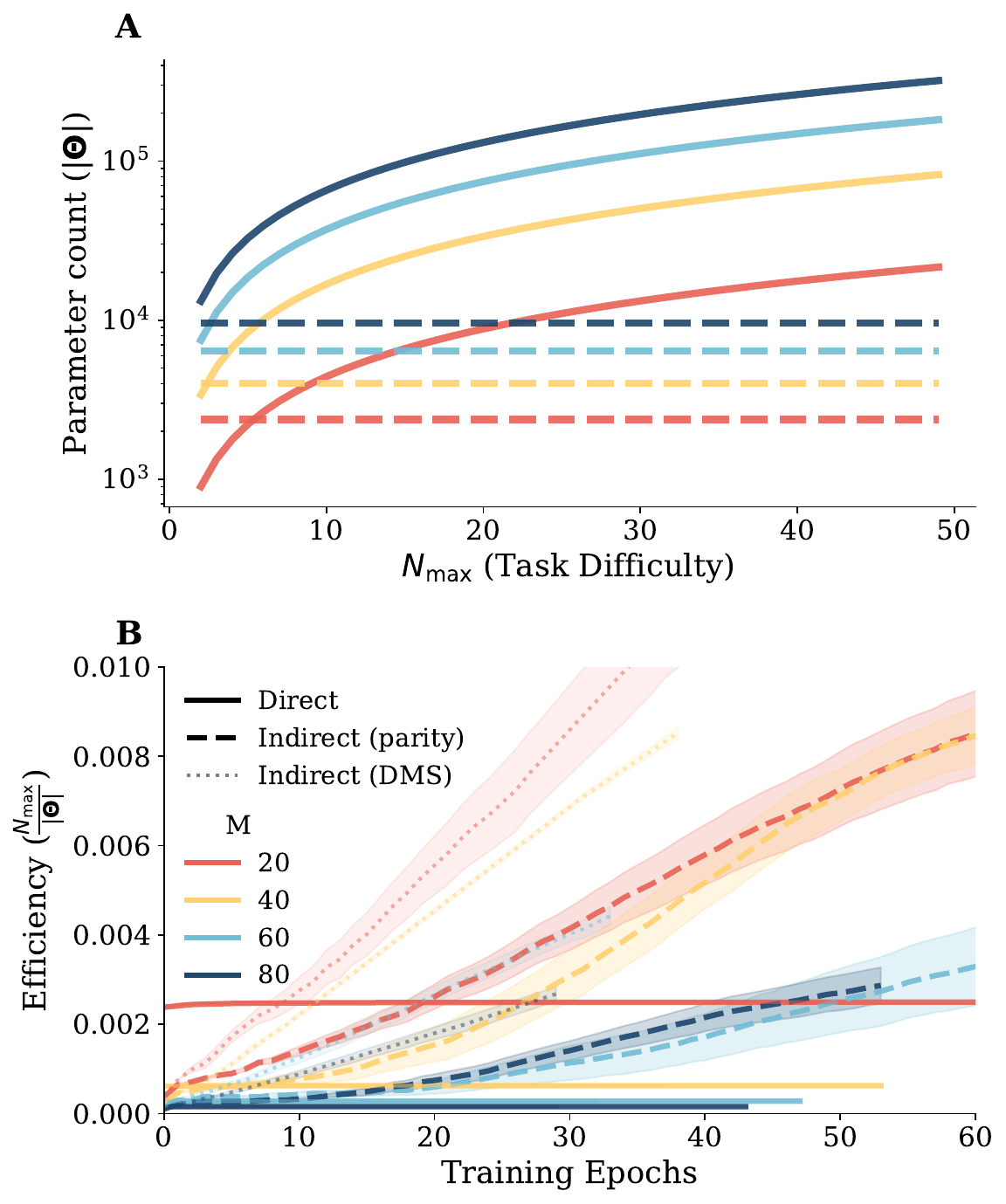}
    \caption{\textit{\textbf{Performance and parameter efficiency. }\textbf{A.} Parameter count ($|\mathbf{\Theta}|$) scaling: directly trained parameters (solid) grow with $N$ and $M^2$, while indirectly trained parameters (dashed) scale much more favorably. \textbf{B.} Learning efficiency (maximum $N$ solved per trainable parameter, $\frac{N_{\max}}{|\mathbf{\Theta}|}$) over training epochs. Solid lines = direct (both tasks), dashed = indirect (parity), dotted = indirect (DMS). The indirect advantage is consistent across both tasks; DMS networks advance faster through the curriculum, consistent with its lower difficulty. Shaded regions indicate SEM across 8 runs per condition.}}
    \label{fig:performance}
    \vspace{-1em}
\end{figure}

\subsection{The g-net generates feedforward connectivity more efficiently} %

Here we show that using a g-net to indirectly learn to generate the connections of the p-net provides a more efficient means to create networks capable of solving equally difficult tasks than a p-net whose weights were all trained directly.

Figure~\ref{fig:performance}A (solid lines) shows that having to learn $\mathbf{W}^{F}_m$ directly levies a heavy cost in terms of number of trainable parameters, $|\mathbf{\Theta}|$. Specifically, $|\mathbf{\Theta}|$ scales linearly with task complexity ($N$) and quadratically with module size ($M$): $|\mathbf{\Theta}| \propto NM^2$. By comparison, the indirect approach using a small g-net (Fig.~\ref{fig:performance}A; dashed lines) to generate $\mathbf{W}^{F}_m$ has much more favorable scaling, with i) the number of parameters remaining nearly constant as a function of task complexity and ii) scaling more favorably with module size $M$.  Thus, a generative function that produces functional connectivity in the form of the hypernetwork allows for a much more efficient scaling towards higher $N$s than direct training would have allowed. The same efficiency advantage is observed for networks trained on the $N$-DMS task (Fig.~\ref{fig:performance}B, dotted lines), confirming that this result is not task-specific. DMS networks advance faster through the curriculum, consistent with its lower difficulty.

Here, the parameter counts are shown for g-nets that are 3 layers deep and 16 neurons wide, and 10 embedding features allocated to each input index, since they proved effective in solving both tasks. %

Next, we define learning efficiency as the complexity of the task solved divided by the number of parameters being trained. Figure~\ref{fig:performance}B compares the relative efficiency of the g-net approach to direct training of the weights over the course of training. The exact epoch at which the g-nets outperform the direct training regime varies depending on the module sizes that are chosen. However, what is unequivocally true is that larger module sizes put the direct learning protocol at a severe disadvantage. The limited efficiency advantage of g-nets at smaller module sizes might, at first glance, diminish the utility of this approach. However, another desirable but orthogonal performance metric is robustness to perturbations, which we explore next to further demonstrate the advantages of the indirect learning strategy via a g-net.

\subsection{Robustness scales with module size}

\begin{figure}[t]
    \centering
\includegraphics[width=0.99\columnwidth]{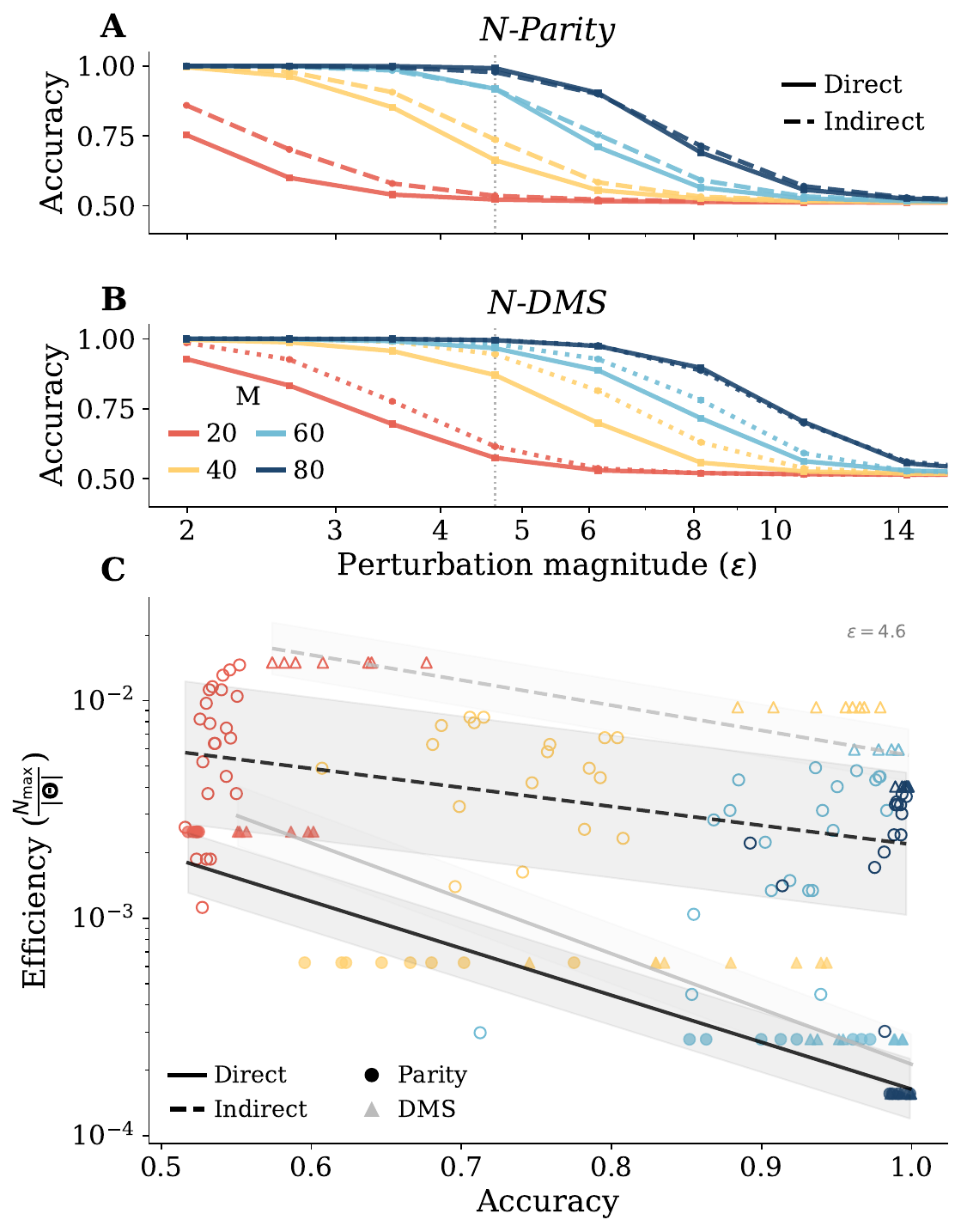}
    \caption{\textit{\textbf{Network robustness and parameter efficiency.} \textbf{A/B.} Accuracy under perturbation of $\mathbf{W}^F$ for parity (A) and DMS (B). Directly trained networks (solid) vs.\ indirectly trained via g-net (dashed); indirectly trained networks are more robust, especially at smaller $M$. \textbf{C.} Efficiency--robustness trade-off at $\varepsilon = 4.6$ (dotted line in A/B). Circles = parity, triangles = DMS; filled = directly trained, unfilled = indirectly trained via g-net. Solid regression = direct, dashed = indirect. The indirect approach offers a better trade-off on both tasks.}}
    \label{fig3_pert_pareto}
    \vspace{-2em}
\end{figure}

Figure~\ref{fig3_pert_pareto}A,B shows how hierarchical networks' performance is affected by perturbations of their connectivity weights for both parity and DMS, where we measured the average accuracy with which the perturbed networks continued to solve the tasks. We perturbed the learned $\mathbf{W}^{F}_m$ connections of both directly and indirectly trained networks that had learned to solve up to and including $N=40$. Different magnitudes of perturbations are indicated by $\varepsilon$, and the effect of these perturbations to the networks' performance.

Networks generated with both training schemes become more robust as the module size increases, as expected from the greater
  redundancy in larger modules. Additionally, at lower module sizes ($M < 60$), g-net generated weights are significantly more robust to the same perturbation magnitude, indicating that the indirect approach confers a functional advantage precisely where networks are most vulnerable.

The performance of these networks can therefore be considered across two dimensions: training efficiency and robustness to noise. Figure~\ref{fig3_pert_pareto}C plots these against each other at $\varepsilon=4.6$ ($N=40$), where each point represents a single trained network. The negative regression slope indicates an inherent trade-off between the two metrics. Despite higher variance, the g-net offers a better trade-off on both tasks, with the Pareto advantage being even more pronounced on DMS.

\subsection{The connectivity  generated by the g-net is more compressible}

\begin{figure*}[t]
    \centering
\includegraphics[width=0.99\textwidth]{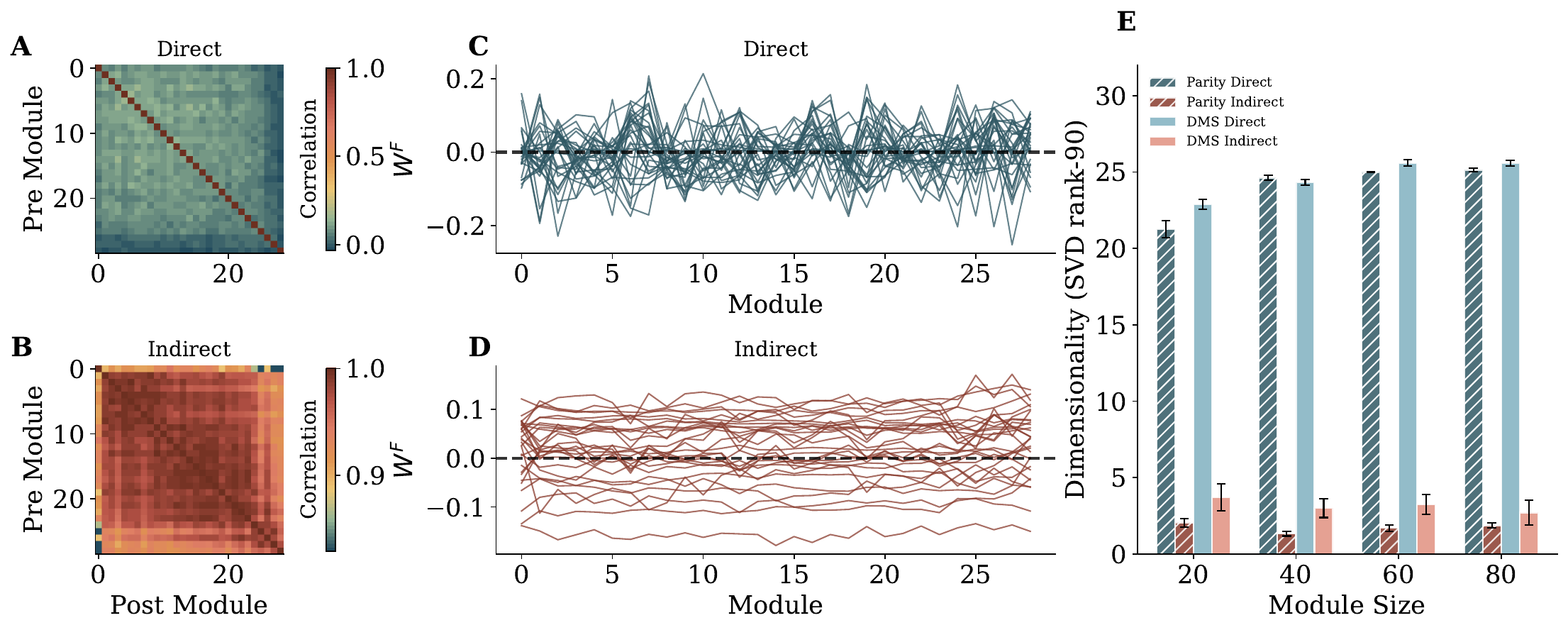}
    \caption{\textit{\textbf{Compressibility of $\mathbf{W}^F$ across modules.} Panels A--D show representative parity networks. \textbf{A.} Directly trained connections are fully uncorrelated between modules, while indirectly learned connections (\textbf{B}) are highly conserved. \textbf{C.} Directly trained weights change rapidly across modules; \textbf{D.} indirectly learned weights show minimal variation. \textbf{E.} SVD rank-90 (number of components for $90\%$ reconstruction fidelity) for both tasks: hatched bars = parity, solid bars = DMS. Indirectly learned connectivity requires only 2--4 components regardless of task or module size, compared to 21--26 for directly trained weights. Error bars indicate SEM.}}
    \label{fig:svd_across}
    \vspace{-1em}
\end{figure*}

\subsubsection{Across-Module Redundancy} We have shown that a generative model producing viable weights can be learned with fewer trainable parameters than when training the weights directly. Thus, one can say that the g-net offers a more compressed way to \textit{encode} the weights. However, one can still ask whether the weights in their fully \textit{decoded} realization are also more compressible than their directly trained counterparts. 

We first examine the conservation of connectivity across modules, i.e., if the connections between specific pairs of neurons remain similar across the duplicated modules. We quantify this by calculating the Pearson cross-correlation of the weights between every module pair in a given realization of a well-performing network. 
Figures~\ref{fig:svd_across}A and C show results for a representative network from each training modality. For directly trained networks, connectivity is largely uncorrelated between modules (mean $r = 0.05$--$0.14$ across module sizes; Fig.~\ref{fig:svd_across}A), meaning that one module's connectivity is not predictive of another's.
In contrast, indirectly learned connectivity shows strong cross-module conservation (mean $r = 0.67$--$0.90$; Fig.~\ref{fig:svd_across}B,D): weights are largely unchanged across modules, and when they do change, they tend to change in a coordinated manner. 

This structure of the g-net generated weights suggests there is a systematic pattern behind the generative process, which is expected given that the g-net parametrizes a continuous function and the module index, $m$, serves as one of the three inputs to this function. 
To further test this assumption, we used singular value decomposition (SVD) to quantify the relative compressibility of the weights generated by networks of different module sizes. Figure \ref{fig:svd_across}E shows the number of components that are needed to reconstruct the weights with 90\% fidelity for each group. For g-net generated weights, only 2-6 components are necessary to reconstruct the weights of the first 30 modules, while more than 25 are needed to reach this same threshold for directly trained weights. It is also worth noting that increasing the module size raises this number (slightly but significantly) for directly learned weights, while remaining the same in the g-net generated scenario. On the DMS task, we observe a similar pattern: g-net generated weights require only $\sim$3 SVD components for 90\% reconstruction fidelity, compared to 23--26 for directly trained weights.

\subsubsection{Within-Module Redundancy} 

\begin{figure}[!htb]
  \centering
\includegraphics[width=0.95\columnwidth]{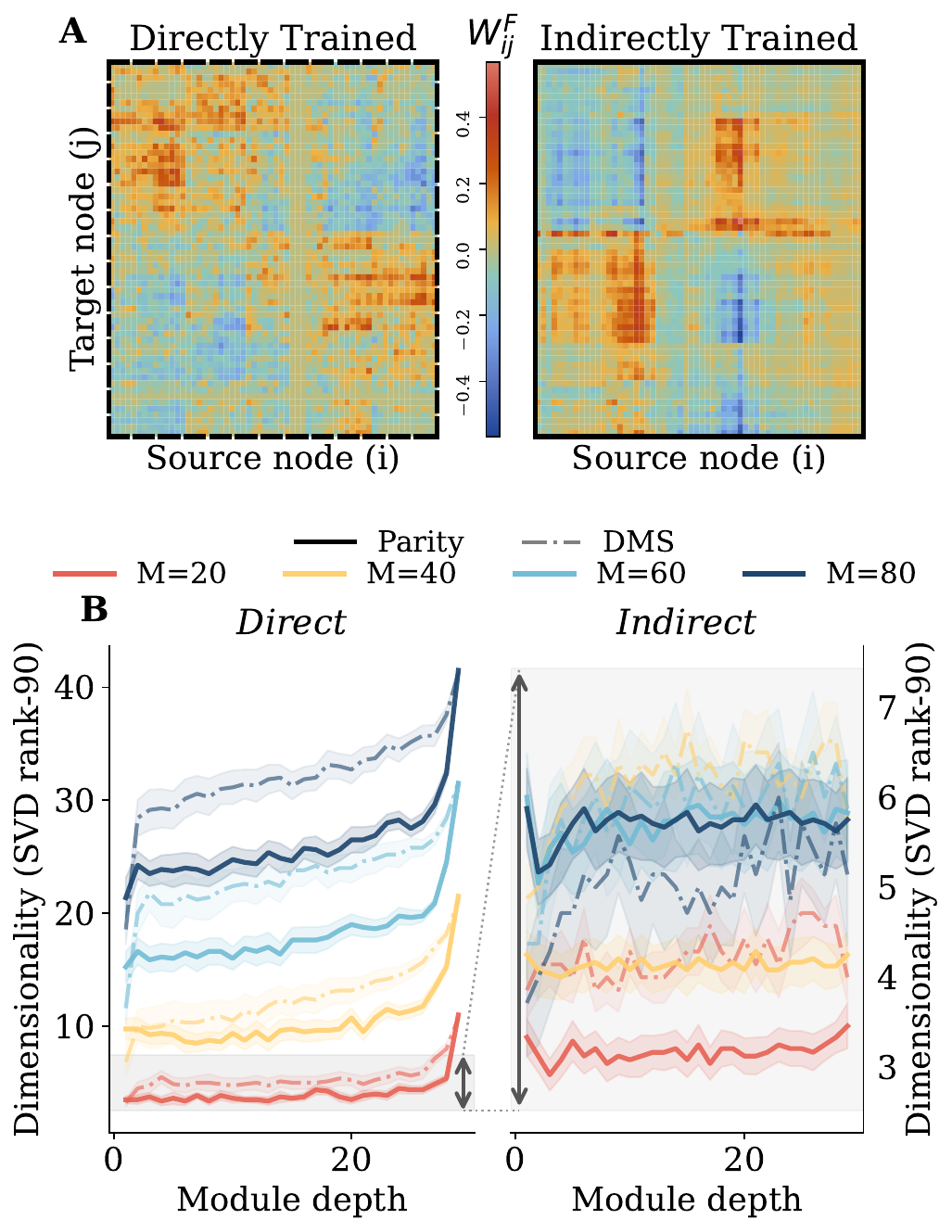}
  \caption{\textit{\textbf{Within-module compressibility of $\mathbf{W}^{F}_m$.} \textbf{A.} Hierarchically clustered weight
  heatmaps at module $m{=}20$ for a representative parity network (dashed border = direct, solid = indirect). Indirectly learned weights exhibit more regular block structure. \textbf{B.} Within-module SVD rank-90 across depth, split by training method: Direct (left) and Indirect (right, note different y-scale). Solid lines = parity, dash-dot = DMS. Directly trained DMS weights have consistently higher rank than parity, while indirectly learned weights remain equally compressed regardless of task. Connecting lines indicate the zoomed range. Shaded regions indicate SEM.}}
    \label{fig:svd_within}
    \vspace{-1em}
\end{figure}

So far, we have examined how the g-net generates highly compressible connections with high redundancy \textit{across} multiple modules. This means that the connection from node $i$ to node $j$ for different $m$ are highly conserved. However, we have not yet examined the compressibility of a connections \textit{within} a single bundle of feedforward weights feeding into a given module. Here we compare the \textit{within-layer} complexity of $\mathbf{W}^{F}_{(m)}$ at different modules ($m$) and for different module sizes ($M$) and training protocols.

Figure~\ref{fig:svd_within}A shows a heatmap of $W^{F}_{ij}$ at module $m=20$ from a representative network trained using each of the two learning strategies. 
The weights were hierarchically clustered to visually reveal the relatively higher redundancy in the g-net generated weights. For a more quantitative comparison, we again use SVD analysis to capture the compressibility of the weights, shown in Figure \ref{fig:svd_within}B. 
Echoing the across-module results, we see that g-net generated weights can be reconstructed using only 3-7 components irrespective of the module size. Note, for comparison, that the maximum rank of these $M \times M$ matrices is $M$, and empirically a random Gaussian matrix requires $\sim M/2$ components for 90\% reconstruction. Thus, reconstructing a weight matrix with $M^2 = 6{,}400$ entries at 90\% fidelity using only 7 out of 80 possible components indicates a great degree of compression. On the other hand, directly trained weights do not benefit from this scaling property and overfit to an expanded $\mathbf{W}^{F}_m$ which is not as easily compressible. 

One final observation from Figure~\ref{fig:svd_within}B involves the relative stability of weight compressibility across depth. For the \textit{directly}-learned weights, modules that correspond to the more recently learned tasks (in this case $N=28-30$) exhibit a sharp increase in their SVD rank, indicating that these weights are less compressible. This is expected: recently added modules have had the least training time, so their directly learned connectivity has not yet converged to a low-rank structure. 
In contrast, the g-net generated weights maintain stable, low-rank structure across all depths, because every module's connectivity is produced through the same compressed bottleneck regardless of when it was added to the curriculum. We interpret this stability as a regularizing effect of the genomic bottleneck \citep{shuvaev2024encoding}. The within-module compression advantage and the stability of layerwise rank across depth are both replicated on the DMS task (Fig.~\ref{fig:svd_within}B), further supporting the task-independence of this regularizing effect. Interestingly, directly trained DMS weights exhibit consistently higher within-module complexity than parity weights across all module sizes and depths (Fig.~\ref{fig:svd_within}B, left panel), with the gap scaling proportionally with overall rank ($\sim$20--30\% higher for DMS).
This is somewhat counterintuitive given that DMS is the simpler of the two tasks. Moreover, while parity weights at early (well-trained) modules converge to relatively low rank, DMS weights maintain elevated rank even at early modules, suggesting that the connectivity patterns required by DMS are inherently less compressible within each module.
In contrast, the g-net produces equally compressed connectivity for both tasks (Fig.~\ref{fig:svd_within}B, right panel), suggesting that the regularization induced by the genomic bottleneck is task-agnostic.

This complements the earlier result from Figure~\ref{fig:performance}A: whereas there we showed the \textit{generative model} (g-net) is more compressible than directly learned weights, here we show its \textit{output} (the generated weights) is also more compressible.

\section{Discussion}
We have shown that combining a hierarchical modular reservoir \citep{Hamidi2024-hx} with a hypernetwork
  that generates inter-module connectivity \citep{Ha2016-vz, shuvaev2024encoding} yields networks that are more
  parameter-efficient, more robust to perturbation, and substantially more compressible than their directly
  trained counterparts.

In addition to solving tasks of equal complexity more efficiently, these networks also scale much more favorably with both task complexity and module size. This feature allows for generating feedforward connections that are significantly more robust to noise, while requiring comparatively fewer parameters to learn. Finally, we show that the dimensionality of the weights generated by the hypernetwork are significantly lower, demonstrating the regularizing effect of the genomic bottleneck in finding equally performative, but massively more compressed solutions than those learned in the direct training scheme. Critically, all reported advantages hold for both parity and DMS (Figures~\ref{fig:performance}--\ref{fig:svd_within}), suggesting that the benefits of the genomic bottleneck are not task-specific but reflect general properties of the indirect encoding strategy.
These results suggest that network size and complexity may be partly dissociated from task difficulty. That a simple generative network can produce highly compressible functional connectivity implies that even strict information bottlenecks need not impair performance, while potentially conferring functional advantages such as robustness.

One unexpected finding is that directly trained DMS weights are less compressible \textit{within} each module than parity weights (Fig.~\ref{fig:svd_within}B, left), despite DMS being simpler. Parity applies the same compositional operation (cumulative XOR) at every depth, which may allow each module's connectivity to converge to a similar low-rank structure, whereas DMS requires delay-specific comparisons that demand richer routing patterns in each $\mathbf{W}^{F}_m$. The slightly higher across-module rank for DMS (Fig.~\ref{fig:svd_across}E) is consistent with this interpretation.

\subsubsection{Relevance to Biological Learning} Evolution compresses generations worth of trial and error learning into the genome. Embryogenesis then decodes this heavily compressed information into a highly developed organism that comes into the world already equipped with functional capabilities that would have otherwise required precious time to learn. %
However, there are other advantages to compressing information beyond this amortization of ancestral experience. One is in enabling evolution to take advantage of the discrete, compositional nature of genes to create systematic variation via duplication, recombination, relaxed selection, exaptation and other means \citep{Czegel2021-qs, Deacon2010-of, Stephenson-Jones2011-qy}. The other is to facilitate the high-fidelity preservation and inheritance of successful variants thanks to the advanced error-correcting mechanisms that a genomic encoding affords \citep{sancar2004molecular, battail2019error}. 

We used backpropagation to learn the parameters of the g-net as well as the p-net. While gradient descent may sufficiently capture some aspects of evolutionary learning \citep{Whitelam2020-sf, da2020natural}, we could use explicit evolutionary learning algorithms in future work to optimize the weights of the g-net. This would allow for the g-net to benefit from additional evolutionary mechanisms such as sexual recombination, and amplify its meta-learning efficacy through the Baldwin effect \citep{Fernando2018-cf}, while simultaneously serving as a more faithful model of the biological processes that inspired it.

\subsubsection{Relevance for Machine Learning}

From a theoretical standpoint, compression induced via information bottlenecks are thought to encourage learning more disentangled representations that lend themselves better to composition \citep{higgins2017beta, Andreas2019-of}. This in turn is argued to facilitate more systematic exploration of solution spaces and better generalization \citep{Laversanne-Finot2018-af, Islam2022-ct, Kim2021-ij}. 

Information bottlenecks are thought to facilitate learning of such representations by encouraging them to model the invariants of the world \citep{Burgess2018-hc, Higgins2018-lm}, and regularize the model away from overfitting to superfluous details in the data. Indeed, ``world models`` \citep{ha2018world} make use of the compressed representations learned by VAEs as inputs to their RNN module which temporally integrates this information. Thus, our genomic bottleneck may induce similar forms of regularization that encourages the RNN to learn reusable \textit{dynamical invariants} for efficiently solving challenging tasks \citep{Shalev-Shwartz2017-uh}. However, investigating this hypothesis will require mechanistic interpretation of our networks which we defer to future work.

From a practical standpoint, memory and compute limitations are presenting hard economic limits to training very large models. Yet, we know that ANNs are dramatically under-trained \citep{hoffmann2022training}, and that much of the performance of an ANN can even be attributed to an isolated \textit{untrained} subset of its parameters \citep{Frankle2018-mv, Malach2020-gu}. %
These results point to the lack of efficiency in training ANNs. Many attempts have been made to lower this deficit by compressing the number of parameters using teacher-student networks \citep{Liu2018-nd, Malik2020-ye}, low-rank approximations  \citep[LoRA][]{hu2022lora}, hypernetworks \citep{Lorraine2018-rg}, response functions \citep{MacKay2019-yw, Bae2022-jy} and other bespoke solutions \citep{houlsby2019parameter, Cheung2019-mj}. 
Our work complements this literature by demonstrating that the combination of hypernetworks, modular architecture, and reservoir computing can provide an additional powerful tool towards more efficient networks.

\subsubsection{Limitations and Future Directions}

Sensitivity to initial conditions is a well known limitation to training hypernetworks \citep{chauhan2024brief}, which we also encountered in our model. Whereas some g-nets are quick to begin to solve tasks and proceed through the curriculum, others, with different random initializations may experience a delay before they enter the curriculum advancing regime. In comparison, learning rates of networks with directly trained weights are much less sensitive to random initializations.   
In practice, we did not feel the need to resort to any alternative initialization mechanisms \citep[e.g.,][]{chang2023} to remedy the situation, because most of the g-nets that were initialized with the standard Kaiming initialization scheme \citep{he2015delving} overcame the early lag within a reasonable number of epochs.

As mentioned earlier, a major focus of future direction will be to shift away from end-to-end backpropagation towards more neuroevolutionary learning procedures \citep{najarro2023towards, nisioti2024growing, Najarro2023selfassembly, pedersen2024structurally, plantec2024evolving, Akiba2024-wm, stanley2019designing} whose proficiency at architecture-search may help us move beyond the hierarchical topology that we have constrained ourselves to so far.

In summary, inspired by the role of genetically encoded mechanisms in pre-wiring brain connectivity, we demonstrate how a hypernetwork can distill the compressed connectivity of a modular reservoir, yielding a robust network that is efficiently trained by modifying only its output connections.

\section{Acknowledgments} %
This work was supported by the German Federal Ministry of Education and Research (BMBF) and the Deutsche Forschungsgemeinschaft (DFG, German Research Foundation) under Germany’s Excellence Strategy. CMW is supported by the European Research Council (ERC) under the European Union’s Horizon 2020 research and innovation programme ($C^4$: 101164709), the Hessian research funding programme LOEWE/4b//519/05/01.002(0022)/119, the Deutsche Forschungsgemeinschaft (German Research Foundation, DFG) under Germany’s Excellence Strategy (EXC 3066/1 ``The Adaptive Mind'', Project No. 533717223), and the Excellence Cluster ``Reasonable AI'' by the Deutsche Forschungsgemeinschaft (German Research Foundation, DFG) under Germany’s Excellence Strategy – EXC-3057.

Claude Code (Anthropic) was used for copy editing and code development during the preparation of this manuscript.

\footnotesize
\bibliographystyle{apalike}
\bibliography{2025}

\end{document}